\pdfoutput=1

\documentclass[11pt]{article}

\usepackage[final]{acl}

\usepackage{times}
\usepackage{latexsym}

\usepackage[T1]{fontenc}

\usepackage[utf8]{inputenc}

\usepackage{microtype}

\usepackage{inconsolata}

\usepackage{graphicx}

\usepackage[utf8]{inputenc} 
\usepackage[T1]{fontenc}    
\usepackage{hyperref}       
\usepackage{url}            
\usepackage{booktabs}       
\usepackage{amsfonts}       
\usepackage{nicefrac}       
\usepackage{microtype}      
\usepackage{xcolor}         
\usepackage{multirow,multicol}
\usepackage{graphicx}
\usepackage{arydshln}
\usepackage{makecell}

\usepackage{enumitem}

\NewDocumentCommand{\gy}{ mO{} }{\textcolor{cyan}{\textsuperscript{\textit{Yi}}\textsf{\textbf{\small[#1]}}}}

\title{Text Grafting: Near-Distribution Weak Supervision for Minority Classes\\in Text Classification}

\author{
  Letian Peng, Yi Gu, Chengyu Dong, Zihan Wang, 
  Jingbo Shang\thanks{$\ $  Corresponding author.}\\
  Department of Computer Science\\
  University of California, San Diego\\
  \texttt{\{lepeng, yig025, cdong, ziw224,
  jshang\}@ucsd.edu} \\
}

\usepackage{xspace}

\usepackage{soul}

\begin{document}
\maketitle
\begin{abstract}
    For extremely weak-supervised text classification, pioneer research generates pseudo labels by mining texts similar to the class names from the raw corpus, which may end up with very limited or even no samples for the minority classes.
Recent works have started to generate the relevant texts by prompting LLMs using the class names or definitions; however, there is a high risk that LLMs cannot generate in-distribution (i.e., similar to the corpus where the text classifier will be applied) data, leading to ungeneralizable classifiers. 
In this paper, we combine the advantages of these two approaches and propose to bridge the gap via a novel framework, \emph{text grafting}, which aims to obtain clean and near-distribution weak supervision for minority classes. 
Specifically, we first use LLM-based logits to mine masked templates from the raw corpus, which have a high potential for data synthesis into the target minority class.
Then, the templates are filled by state-of-the-art LLMs to synthesize near-distribution texts falling into minority classes.
Text grafting shows significant improvement over direct mining or synthesis on minority classes. 
We also use analysis and case studies to comprehend the property of text grafting.


\end{abstract}

\section{Introduction}

Recent research has made rapid progress on extremely weak-supervised text classification (XWS-TC)~\cite{x-tc}, limiting the supervision to a brief natural-language description without any annotated samples. 
For example, text mining-based XWS-TC \citep{lot-class,x-class,taxo-class,lops,npprompt,debiase-sota} takes only class names or seed words from humans and discovers potential in-class texts following designated heuristics. 

Minority classes are arguably the most challenging part of XWS-TC.
The class distribution in real-world datasets is often a long-tailed distribution~\cite{longtailed}, with a non-trivial number of minority classes.
These minority classes have a very small number of documents in the raw corpus, therefore, it is difficult to locate the right documents by mining-based methods, leading to noisy pseudo-labels.
Under extreme circumstances, the mining-based methods may end up with no sample for minority classes.

\begin{figure}[t]
    \centering
    \includegraphics[width=\linewidth]{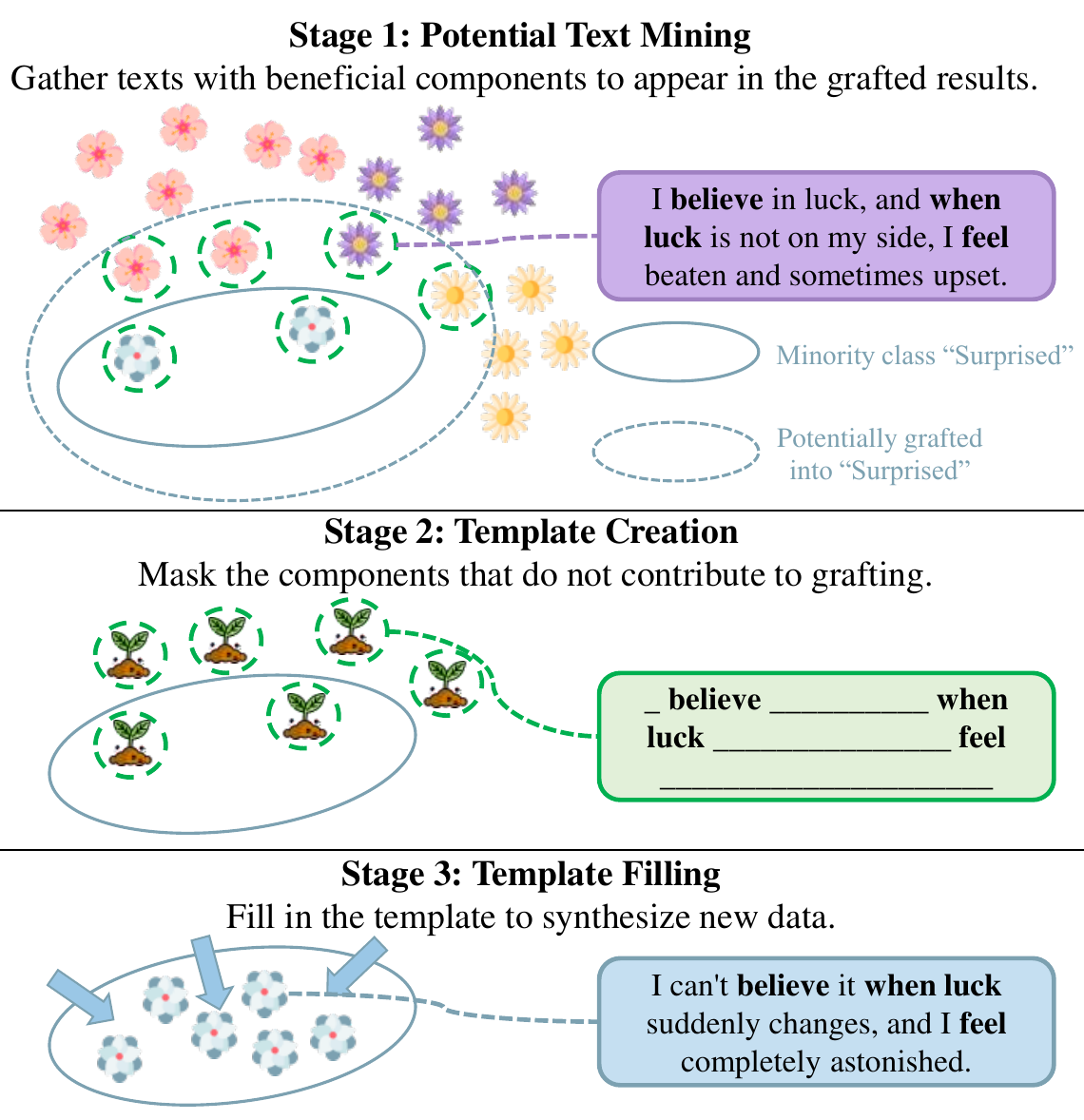}
    \vspace{-8mm}
    \caption{The framework of text grafting.
    }
    \label{fig:show}
\end{figure}

\begin{table}[t]
\small
\centering
\resizebox{\linewidth}{!}{
    \begin{tabular}{lcccc}
\toprule
\thead{Framework} & \thead{Mining?} & \thead{Train Data} & \thead{Data Quality} & \thead{In-Distribution} \\
\midrule
Text Mining & Text & Raw & \textcolor{red}{Noisy} & \textcolor{green}{Yes} \\
Data Synthesis & None & Generated & \textcolor{green}{Clean} & \textcolor{red}{Hardly} \\
\midrule
\textbf{Text Grafting (ours)} & Template & Grafted & \textcolor{green}{Clean} & \textcolor{green}{Mostly} \\
\bottomrule
\end{tabular}
}
\caption{High-level comparison among three discussed XWS-TC frameworks.} 
\vspace{-5mm}
\label{tab:comparison}
\end{table}

A potential way to address this issue is data synthesis-based XWS-TC~\cite{zerogen,progen,incubator}, which hopes to generate in-class texts by prompting large language models (LLM)~\citep{gpt-3,gpt-4,llama,llama2,llama-3,gemma,openai_hello_gpt4o} with class names or definitions. 
However, such synthesized texts may follow a distribution different from the corpus where the text classifier will be later applied~\citep{detectgpt}, which makes the learned text classifier out-of-distribution, leading to poor performance.

This paper combines the advantages of mining-based and synthesis-based frameworks to propose a new framework, \emph{text grafting}, which aims to obtain clean and near-distribution weak supervision for minority classes. 
As specified in Figure~\ref{fig:show}, text grafting incorporates three stages: (1) \emph{Potential Text Mining} gathers raw texts with beneficial components to synthesize in-class texts for the target minority class. (2) \emph{Template Creation} forms templates by masking the components that do not contribute to the in-class text synthesis. (3) \emph{Template Filling} synthesizes in-class texts by filling in the blanks. 
Table~\ref{tab:comparison} systematically compares the weak supervision obtained by different frameworks. 


To identify the words not contributing to the classification, we borrow the marginalization idea from LLM reasoning \cite{dcpmi}. 
We get the probability logit of each word in the raw text by instructing LLMs (relatively small, specifically Gemma~\cite{gemma}) to generate with or without the in-class as a requirement. 
The difference between the two logits represents the potential of each word to appear in the grafted text.
As only words with high potential will be left, we use the average potential of top-$K\%$ words to represent the text potential score. The bottom-$(100-K)\%$ words will be masked to form the template for data synthesis. We rank the templates by their potential scores and select top-$T\%$ templates for the last template-filling stage. Finally, these selected templates are filled by prompting a state-of-the-art LLM, GPT-4o~\citep{openai_hello_gpt4o}. 

We compare the three mentioned frameworks on various raw corpora to classify different minority classes. 
The experiment results show text grafting can outperform state-of-the-art text mining and dataset synthesis methods. 
The ablation study verifies that all stages and the intermediate template contribute to the success of our proposed text grafting. The mask-and-filling scenario also shows its advantage over simple in-context generation, since it forces the LLM to incorporate components from the raw texts. We also involve an extreme situation where the target class does not appear in the raw corpus {completely}. 
Remarkably, text grafting shows its robustness to this extreme {situation}, indicating its {applicability} does not require the target class to appear in the raw corpus. 
This enables text grafting to work on a very small corpus {which} boosts efficiency. 

Furthermore, we analyze and discuss the property of text grafting. We apply principal component analysis to visualize that the drafted texts are indeed near in-distribution. We also find the grafted texts are near-distribution enough that we do not need to synthesize negative samples as in traditional data synthesis, which reduces the cost. {We also conduct a comprehensive hyperparameter analysis of our method. Interestingly, we found that} The mask ratio is searched to be better set to a high value like $0.75$ and the mined template number can be as small as $200$. These case studies explore the advantages of text grafting in distribution approximation and its failure when the raw texts are near the distribution of LLM generation. 

We summarize our contributions as follows,
\begin{itemize}[nosep,leftmargin=*]
    \item We propose a novel XWS-TC framework for minority classes, text grafting, combining the in-distribution advantage of text mining and the in-class advantage of data synthesis.
    \item We implement text grafting following the marginalization idea from LLM reasoning, utilizing the probability logits for template mining and masking.
    \item We provide comprehensive analysis and case studies to show the strength, property, and possible failure of text grafting.\footnote{The datasets and models used in the experiments are released in \href{https://github.com/KomeijiForce/TextGrafting}{github.com/KomeijiForce/TextGrafting}}
\end{itemize}

\section{Related Works}

Extremely Weak-Supervised Text Classification (XWS-TC) needs only minimal human guidance to label the text, such as a few rules by human experts that match the text to the labels~\cite{x-tc}. Mainstream XWS-TC methods can be divided into two categories: \textbf{Text Mining} and \textbf{Data Synthesis}. 

\paragraph{Text Mining} is a fundamentak task~\citep{data_mining} for natural language processing. In XWS-TC, the text miner follows high-level rules from humans to annotate raw texts, which are used to train the text classifier. A mainstream rule is whether a seed word appears in the raw text~\cite{contextualized_seed_word,lot-class,x-class}, categorized as seed methods. Another mining way is to prompt language models for logits that reflect the probability of texts falling in classes~\cite{gpt-3}, which can be calibrated by several techniques~\cite{dcpmi,calibration,prototypical_calibration}. The strong performance of existing text mining methods is highly dependent on the precision of the class-indicative rules~\cite{debiase-sota}, which is hard to maintain for minority classes.

\paragraph{Data Synthesis} \cite{data_synthesis} addresses the precision degradation in text mining by directly prompting LLMs with the label names to generate in-class texts~\cite{zerogen,incubator}. With the powerful generative ability of LLMs, the synthesized texts are generally clean (in-class) for training strong classifiers. However, synthesized texts hold LLM-specific patterns, discovered by LLM-generated text detectors~\cite{detectgpt,detection_survey}. This pattern is hard to be eliminated even with in-context learning~\cite{icl_detection}. Thus, synthesized texts are generally out-of-domain and consequently fine-tune a weaker classifier on the test set. 

\paragraph{Minority Classes} widely appear in classification datasets as a result of long-tailed distribution~\cite{longtailed, imbalance_tc}. For minority classes with supervised annotations, techniques like re-sampling~\cite{CAS,resampling,resampling_balancing} and data augmentation~\cite{EDA,da_toxic,MISO,ECRT}. However, these methods are applied to unbalanced annotations, which are unavailable under XWS. 

\paragraph{Counterfactual Augmentation} refers to generating annotated data out of the dataset or raw corpus. Different from regular augmentation, counterfactual augmentation changes the reference, e.g., label flipping~\cite{flipda,cotam}. Counterfactual augmentation is also applied for text-to-text tasks like translation~\cite{cag_nmt} or summarization~\cite{cag_sum}. Counterfactual augmentation shares the same requirement for known reference as regular augmentation. This paper explores a counterfactual augmentation method for unannotated raw text under XWS.
\section{Text Grafting}

\subsection{Preliminary}

\paragraph{XWS Minority Class Classification} takes a raw corpus $\mathcal{D} = \{X_{(i)}\}_{i=1:|\mathcal{D}|}$ and the target minority class name $c$ as the input to train a binary classifier $f(X)$ that discerns a text falling in $c$ or not. We denote the $j$-th word in the $i$-th text of the raw corpus as $x_{(i,j)}$.

\paragraph{Text Mining} gathers in-class texts with high-level rules $g(X)$ that can precisely assign $X$ to target class $c$. Example rules include whether $X$ contains words indicating $c$ (seed words)~\cite{debiase-sota} or $X$ has top confidence to be in $c$ by prompting LLMs~\citep{gpt-3} among $\mathcal{D}$. The mined $D^{(TM)} = \{X_{(i)}|g(X_{(i)})\}_{i=1:|\mathcal{D}|}$ is combined with some randomly sampled negative texts (due to the scarcity of $c$) to train $f(\cdot)$.

\begin{figure}
    \centering
    \scalebox{1.0}{\includegraphics[width=\linewidth]{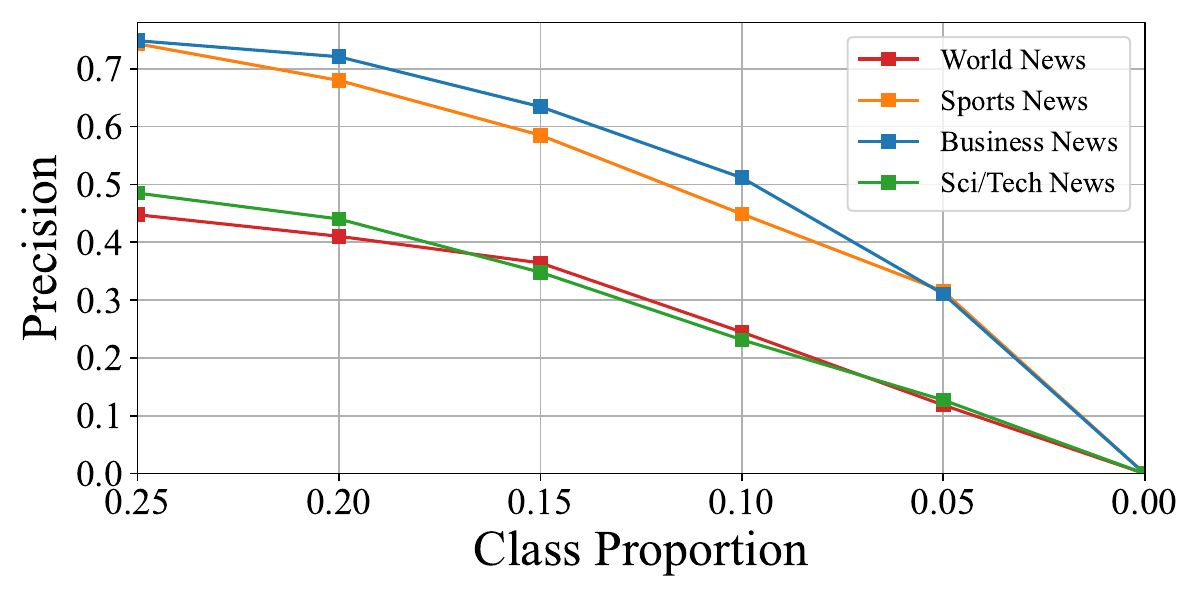}}
    \vspace{-5mm}
    \caption{The precision of state-of-the-art text mining on same classes with different class proportions. ``Precision'' refers to the precision of the pseudo-labels. ``Class Proportion'' means the ratio of the texts of this class in the entire corpus after down-sampling.}
    \vspace{-5mm}
    \label{fig:noise}
\end{figure}

However, text miners fail in minority classes due to their low proportion in the raw corpus. By running a state-of-the-art text mining method~\cite{debiase-sota} on AG-News~\cite{ag_news} with class {name} proportion modified by sampling, we observe the mining precision drops sharply with the decrease of proportion, presented in Figure~\ref{fig:noise}. Another concern is the class might be too minor that even no ground truth can be mined from the raw corpus, limiting the precision to $0\%$ no matter how intuitive the mining rule is.

\paragraph{Data Synthesis} does not annotate raw texts for classifier fine-tuning but directly prompts LLMs to generate in-class texts ($X' \sim \textrm{LLM}(I_c)$), where $I_c$ is an instruction to write a text in class $c$. With the strong capability of state-of-the-art LLMs~\cite{openai_hello_gpt4o,llama-3}, the generated $X'$ are highly confident to fall in class. Another advantage of data synthesis is the ability of LLMs to generate negative samples~\cite{zerogen,incubator}. However, synthesized texts consist of patterns different from other sources~\cite{detectgpt}, which indicates classifiers $f(\cdot)$ fine-tuned by synthesized texts are out-of-domain, consequently weaker in the classification task.

\subsection{Overview of Text Grafting}

\begin{figure*}
    \centering
    \scalebox{1.0}{\includegraphics[width=\linewidth]{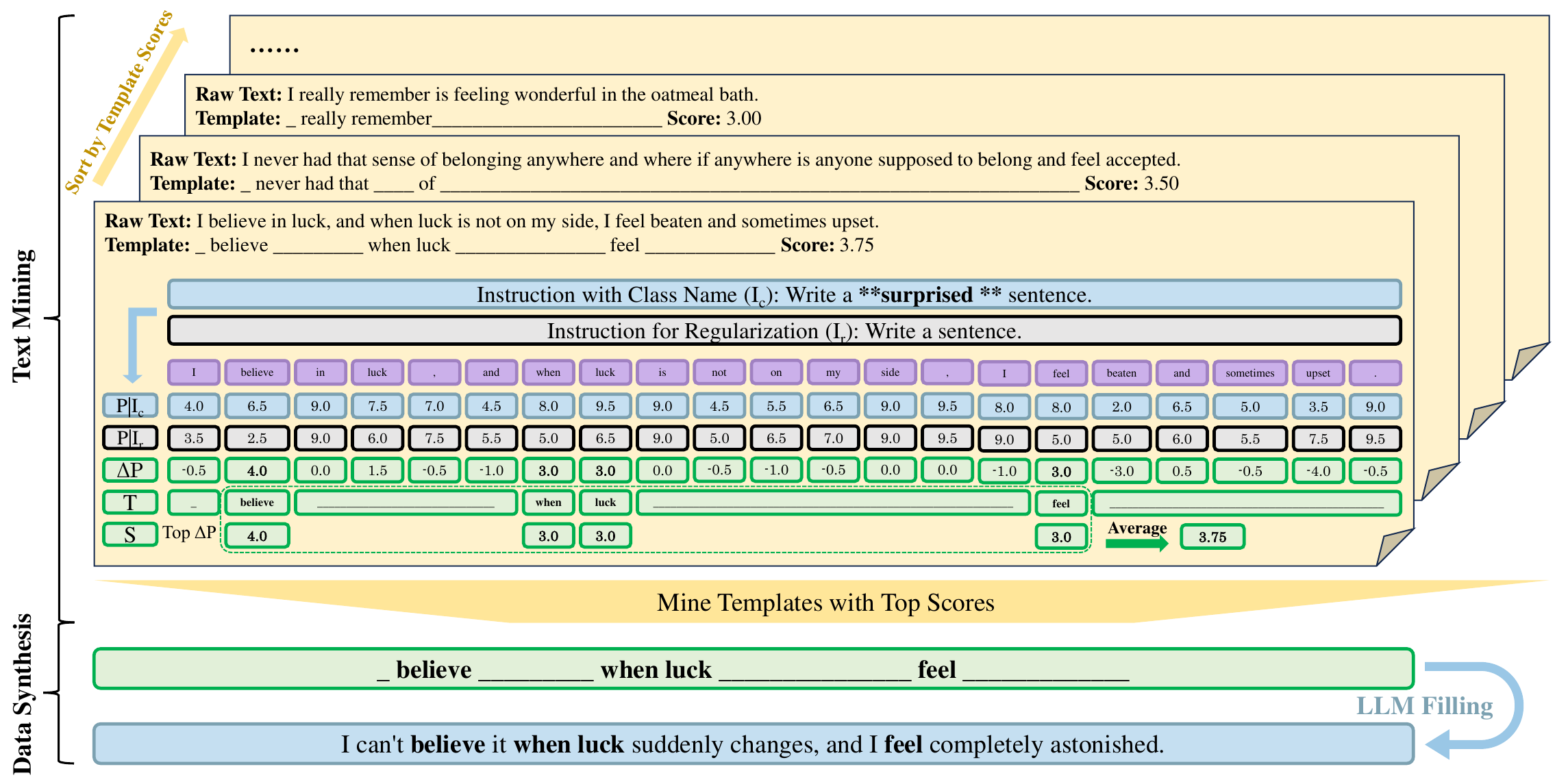}}
    \caption{The overview of text grafting with the minority class \textbf{``Surprised''} in the Emotion dataset as an example. Text grafting includes two stages: \textbf{1) Text (Template) Mining:} Create scored templates and select the ones with the top scores. \textbf{2) Data Synthesis: }Prompt the LLM to fill in the templates to synthesize in-class texts.}
    \label{fig:overview}
\end{figure*}

As depicted in Figure~\ref{fig:overview}, our text grafting is a hybrid method that combines the strengths of text mining and data synthesis. The core observation is that out-of-class texts can contain useful components for writing in-class texts. The text mining stage of text grafting aims to discover these potential components and formalize them as templates. In the data synthesis stage, the templates are filled by LLMs to produce in-class texts. With components from both raw texts and synthesis, the grafted texts are both in-class and near-distribution, which are supposed to fine-tune a better classifier than only text mining or data synthesis. 

\subsection{Implementation}

In detail, the text mining stage includes \textbf{Potential Text Mining} and \textbf{Template Creation}, while in the data synthesis {stage we conduct} \textbf{Template Filling}. The text mining stage requires relatively small open-source LLMs with higher efficiency and accessible logits. Template Filling can utilize state-of-the-art LLMs even with API accessibility. 

\paragraph{Potential Text Mining} discovers texts with potential components to appear in the grafted texts. We evaluate the potential of each word $x_{(i,j)}$ in the raw text $X_{(i)}$ with regularized logits prompted from LLMs following the regularization idea in DC-PMI~\cite{dcpmi}. The potential $\Delta p_{(i,j)}$ for $x_{(i,j)}$ is defined as the difference between the probability logit of $x_{(i,j)}$ prompted by an instruction with the class name ($I_c$) and an instruction for regularization ($I_r$). The difference can also be viewed as the probability of $x_{(i,j)}$ raised by incorporating the class name $c$ into the instruction. 

\vspace{-3mm}

\begin{equation}
    \small
    \Delta p_{(i,j)} = \log P_{\mathrm{LLM}}(x_{(i,j)}|I_c) - \log P_{\mathrm{LLM}}(x_{(i,j)}|I_r)
\end{equation}

The words with top-$K\%$ $\Delta p$ among the words in text $X_i$ will remain in the template. Thus, the average of their $\Delta p$ represents the potential ($\Delta P_i$) of the template created based on $X_i$. As we are mining potential templates rather than directly in-class texts, the mining rate $K\%$ can be much larger than text mining.

\vspace{-5mm}

\def\-{\scalebox{0.5}[1.0]{\( - \)}}
\begin{equation}
    \small
    \Delta P_i = \left \lceil \frac{1}{K\%\cdot|X_i|} \right \rceil \sum_{\Delta p_i \in \mathrm{Top}\- K\%(\Delta p_{1:|X_i|})} \Delta p_i
\end{equation}

Then the texts are ranked by their grafting potential $\Delta P$ and texts with top-$N\%$ potential are mined to create the templates. 

\paragraph{Template Creation} simply masks the words with bottom-$(100-K)\%$ potential $\Delta p$ by blank tokens ``\_'' and uses the top-$K\%$ as template part. Text $X_i$ is thus converted to template $T_i$, which is prepared for LLMs to fill in during the data synthesis stage. As the example in Figure~\ref{fig:overview}, the components with the top potential to be in a grafted ``Surprised'' remain in the template such as ``believe'', ``when luck'', ``feel''. These components support the data synthesis to better write an in-class text while keeping the style in distribution with the writing structure from the raw corpus. 

\paragraph{Template Filling} prompts an LLM to fill in the blanks in $T$, which produces a grafted text that generally falls in the target class $c$. Referring to the example in Figure~\ref{fig:overview}, the LLM well utilizes the writing structure in the template and fills in the blanks to produce the in-class text. As the template keeps the writing structure of the raw corpus, the grafted text is quite similar to the original one but flipped into the target minority class.

\begin{table}
\centering
\small
\scalebox{1.0}{
\begin{tabular}{p{1.5cm}p{5.0cm}}
\toprule
Function & Prompt\\
\midrule
TM ($I_c$) &  ``Please write a <label> <style>.''\\
TM ($I_r$) &  ``Please write a <style>.''\\
\midrule
DS &  ``Fill in the blanks in the template to produce a <label> <style>.''\\
\bottomrule
\end{tabular}
}
\caption{The prompts used in text grafting. In prompts, \textbf{<label>} refers to the label names like ``Surprised'' while \textbf{<style>} represents the distribution like ``Tweet''.} 
\vspace{-5mm}
\label{tab:prompt}
\end{table}

Specific prompts in these stages are shown in Table~\ref{tab:prompt}, where the label and distribution information is filled to support the text grafting. 
\section{Experiments}

\subsection{Evaluation}

\paragraph{Datasets} We take several minority classes from popular text classification datasets to evaluate the performance of different XWS-TC methods on minority classes. We include 1) TweetEval~\cite{tweeteval} and Emotion~\cite{emotion}, which contain minority emotion classes ``Optimism'' ($8.9\%$) and ``Surprised'' ($3.6\%$); 2) 20 News~\cite{20news}, which contains minority news topic ``Religion'' ($3.3\%$) and ``Politics'' ($4.1\%$); 3) BigPatent~\cite{big_patent}, which contains minority patent class ``Mechanical Engineering'' ($7.0\%$). The raw corpus is down-sampled to $10,000$ samples to improve experiment efficiency and save budget costs. We use the F1 score as the metric for evaluation.

\paragraph{Baselines} We include various text mining and data synthesis methods as the baselines for comparison to illustrate the advantage of our text grafting.

Text mining methods include,

\begin{itemize}[nosep,leftmargin=*]
    \item \textbf{Prompting Confidence}~\citep{gpt-3}, which is a prompting method that directly queries an LLM whether the text falls in the target minority class, and uses the probability logit of answering ``yes'' for ranking. Considering the class minority, the mining rate is set to $1\%$.
    \item \textbf{Debiased Seed Word}~\cite{debiase-sota}, which is the current state-of-the-art XWS-TC method. This method uses a seed word (the same as the label name) to match the target minority class and then drops the seed word from the context to eliminate spurious correlation. Then the texts are filtered by text selection~\cite{lops} to produce the final mined texts.
\end{itemize}

Data synthesis methods include,

\begin{itemize}[nosep,leftmargin=*]
    \item \textbf{ZeroGen}~\cite{zerogen}, which directly prompts the LLM to synthesize texts in or out of the target minority class. 
    \item \textbf{In-Context Generation}~\cite{icl-survey}, which uses raw texts as the in-context examples to generate texts with a similar writing style as the raw corpus.
    \item \textbf{Incubator}~\cite{incubator}, which uses instruction-tuned LLMs and in-context learning based on annotated instruction-to-dataset samples to generate data points for fine-tuning. 
\end{itemize}

All text synthesis methods synthesize $1000$ texts as positive (in the target minority class) or negative samples (out of the target minority class, $2000$ in total).

The LLM used for text mining is a popular and advanced open-source LLM, Gemma~\cite{gemma} (\texttt{gemma-1.1-7b-it}) with accessible possibility logits. The LLM used for data synthesis is the state-of-the-art LLM, GPT-4o~\cite{openai_hello_gpt4o}. 

\paragraph{Grafting Hyperparameters} The mining rates of our text grafter are set to $25\%$ ($K\%$) for potential components in templates and $10\%$ ($N\%$) for potential templates. Thus, the synthesized data number is less than $1000$, not more than the data number from pure data synthesis. 

\paragraph{Fine-tuning Hyperparameters} We fine-tune a RoBERTa-Large~\cite{roberta} as the classifier with the AdamW~\cite{adamw} as the optimizer whose learning rate is initialized to $1\times 10^{-5}$. The classifier is fine-tuned by $10$ epochs with batch size $8$ and $20\%$ training data are split for validation to select the best-performing checkpoint. All the experiment results are achieved by an average of $5$ runs. The two stages in text grafting apply the same LLM as text mining and data synthesis.

\newcommand{\specialcell}[2][c]{%
  \begin{tabular}[#1]{@{}c@{}}#2\end{tabular}}
  
\begin{table*}
\centering
\small
\begin{tabular}{llcccccc}
\toprule
\multicolumn{2}{l}{\textbf{Dataset}} & \textbf{TWEET} & \textbf{PATENT} & \textbf{EMOTION} & \multicolumn{2}{c}{\textbf{20NEWS}} & \multirow{4}*{\textbf{Average}} \\
\multicolumn{2}{l}{Distribution} & Tweet & Patent & Tweet & \multicolumn{2}{c}{News} &  \\
\multicolumn{2}{l}{Minority Class} & Optimism & Mechanical & Surprised & Religion & Politics &  \\
\multicolumn{2}{l}{Class Proportion} & $8.9\%$ & $7.0\%$ & $3.6\%$ & $3.3\%$ & $4.1\%$ &  \\
\midrule
\multicolumn{2}{l}{Supervised} & $45.88$ & $34.30$ & $32.28$ & $24.10$ & $32.27$ & $35.14$  \\
\midrule
\midrule
\multirow{2}*{{\specialcell{Text Mining \\(TM)}}}
 & Prompting Confidence & $17.93$ & $14.59$ & $7.00$ & $6.50$ & $15.77$ & $12.81$  \\
 & Debaised Seed Word & $19.15$ & $20.46$ & $8.78$ & $11.47$ & $19.53$ & $15.88$ \\
\midrule
\multirow{3}*{{\specialcell{Data Synthesis \\(DS)}}} 
& ZeroGen & $10.82$ & $24.17$ & $7.19$ & $6.97$ & $17.60$ & $13.35$ \\
& Incubator & $22.46$ & $20.86$ & $7.44$ & $23.96$ & $24.48$ & $19.84$ \\
& In-Context Generation & $16.24$ & $24.53$ & $22.24$ & $21.98$ & $24.13$ & $21.83$ \\
\midrule
\multirow{1}*{TM+DS}
& Text Grafting \textbf{(Ours)} & $\textbf{32.70}$ & $\textbf{25.42}$ & $\textbf{27.46}$ & $\textbf{25.32}$ & $\textbf{27.32}$ & $\textbf{27.64}$ \\
\cmidrule(lr){1-8}
\multirow{4}*{Ablation} & $\quad$w/o Mining & $26.54$ & $16.74$ & $24.32$ & $17.69$ & $15.16$ & $20.09$ \\
& $\quad$w/o Synthesis (DC-PMI) & $17.86$ & $11.34$ & $7.34$ & $4.33$ & $4.28$ & $9.03$  \\
& $\quad$w/ Random Masking & $30.11$ & $19.07$ & $23.37$ & $23.57$ & $26.65$ & $24.55$ \\
& $\quad$w/ MF $\rightarrow$ ICG & $21.31$ & $20.58$ & $15.33$ & $23.60$ & $25.06$ & $21.18$ \\
\midrule
\midrule
\multirow{3}*{Zero-Occur} & Debaised Seed Word & $0.00$ & $17.66$ & $5.88$ & $8.79$ & $20.73$ & $10.61$ \\
& In-Context Generation & $18.84$ & $23.15$ & $19.50$ & $20.63$ & $24.11$ & $21.25$ \\
 & Text Grafting \textbf{(Ours)} & $\textbf{30.61}$ & $\textbf{25.27}$ & $\textbf{31.08}$ & $\textbf{26.15}$ & $\textbf{25.54}$ & $\textbf{27.73}$ \\
\bottomrule
\end{tabular}
\caption{Text mining performance (F1 Score) for minority classes among different datasets.}
\label{tab:main}
\end{table*}

\subsection{Main Result} 

\begin{table}[t]
\small
\centering
\scalebox{1.0}{\begin{tabular}{lcccc}
\toprule
Method & \textbf{EMOTION} & \textbf{TNEWS} \\
Language & English & Chinese \\
\midrule
Debiased Seed Word & $19.14$ & $22.84$ \\
$\quad$ + Text Grafting  & $\textbf{31.30}$ & $\textbf{28.61}$ \\
\bottomrule
\end{tabular}}
\caption{Results (Macro F1 Score) on end-to-end XWS-TC for different languages. Emotion (English) contains minority classes ``Surprised'' and ``Love'' while TNEWS (Chinese) has a minority class ``Stock''.} 
\label{tab:e2e}
\end{table}

The main results from our experiments are presented in Table~\ref{tab:main}. The comparison inside text mining methods shows the advantage of the seed method over the prompt method, consistent with the findings of \citeauthor{x-tc}. The comparison among text synthesis methods reflects the importance of knowledge about the distribution of the corpus, as in-context generation outperforms other baselines with raw texts as an example for synthesis. Finally, text grafting outperforms all the baselines, which verifies the benefit of text grafting to produce in-class and near-distribution texts. 

However, there is still a significant gap between the performance of supervised classification and XWS-TC even with text grafting. This indicates the grafted texts still have differences with the raw corpus distribution for further improvement.

\begin{figure}
    \centering
    \scalebox{0.9}{\includegraphics[width=\linewidth]{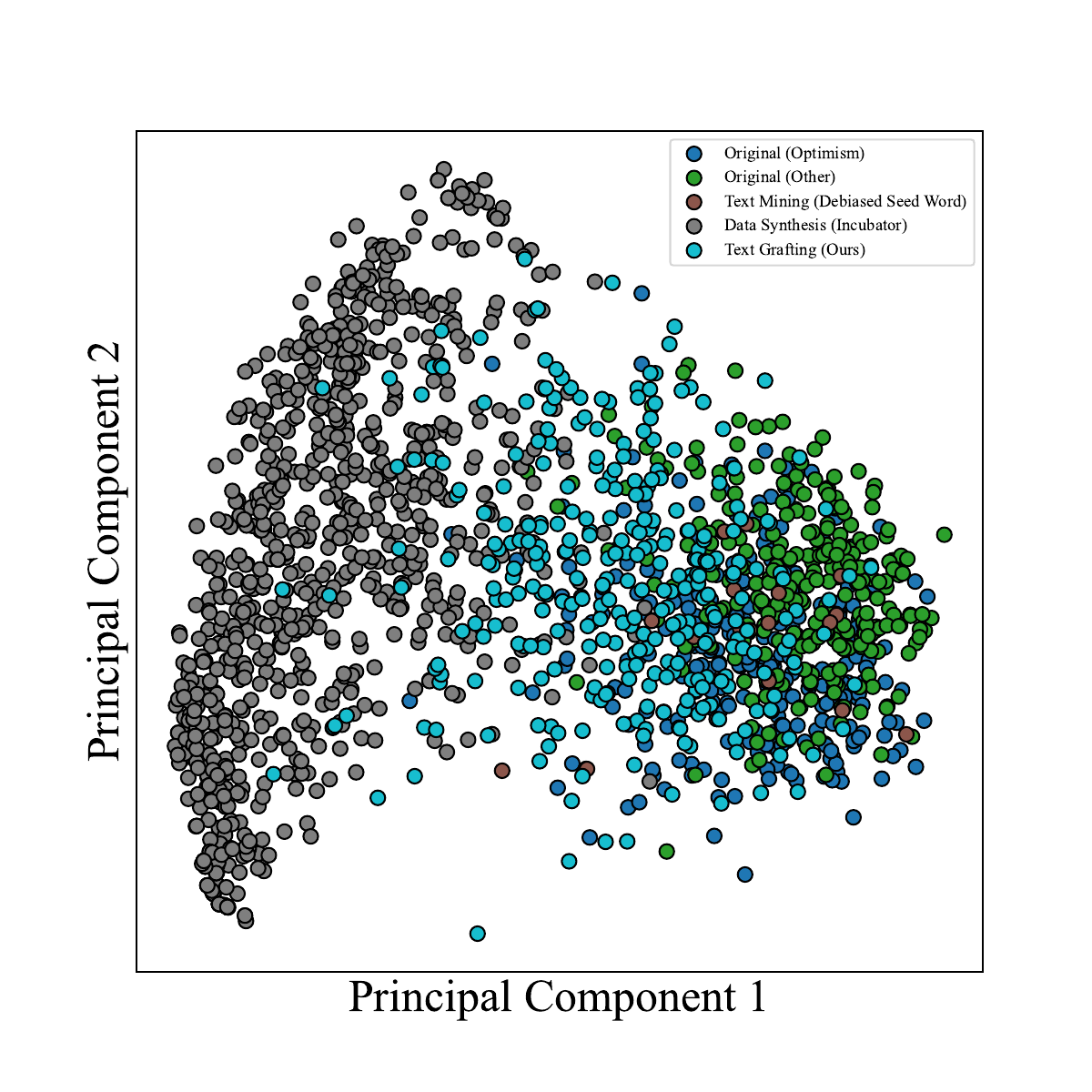}}
    \vspace{-5mm}
    \caption{The visualization of text distributions from different methods.}
    \label{fig:distribution}
\end{figure}

\subsection{Ablation Study} 

Table~\ref{tab:main} also includes the ablation study results for text grafting in the \emph{Ablation} columns. The first comparison focuses on the necessity of text mining and data grafting in the pipelines of text grafting. \textbf{Without Mining} removes the template score-based sorting and lets the LLM fill in randomly selected templates, which significantly underperforms the initial grafting. \textbf{Without Synthesis} does not create templates for data synthesis, but directly uses the $\Delta p$ averaged over all words to mine texts for fine-tuning, equal to DC-PMI~\cite{dcpmi}. The result is similar to the Prompting Confidence method, which shows the limitation of text mining for minority classes. Then we emphasize the necessity of intermediate templates. \textbf{With Random Masking} randomly masks the mined texts instead of following the word-level potential $\Delta p$, which also results in a performance drop. \textbf{With Mask Filling $\rightarrow$ In-Context Generation} takes the mined texts as the in-context examples, which result in a similar performance as the one without mining, indicating the importance of template creation and filling. Based on these ablation results, our grafting framework is shown to be essential for achieving optimal performance by effectively combining data synthesis, text mining, and templates. 

\begin{figure}
    \centering
    \scalebox{1.0}{\includegraphics[width=\linewidth]{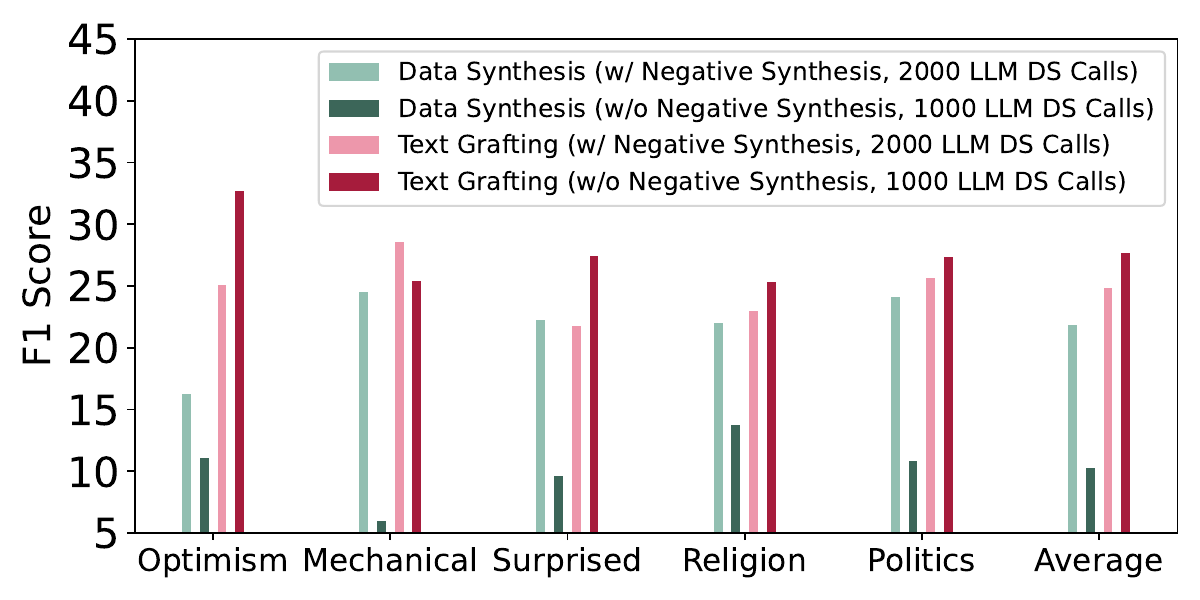}}
    \vspace{-6mm}
    \caption{The analysis on the necessity of negative data synthesis.}
    \label{fig:negative_synthesis}
\end{figure}

\subsection{Further Analysis}

\paragraph{Q1: How does Text Grafting Benefit End-to-End XWS-TC?} 

 Table~\ref{tab:e2e} shows how text grafting can be integrated into end-to-end XWS-TC pipelines for different languages. We include the English Emotion dataset with ``Surprised'' and ``Love'' as the minority classes and the Chinese TNEWS dataset~\cite{clue} with a minority class ``Stock''. For the minority classes, texts are synthesized by grafting while other classes apply the traditional debiased seed word method. The result shows text grafting improves end-to-end XWS-TC on different languages, which verifies the cross-lingual benefit of integrating text grafting into XWS-TC pipelines to handle minority classes. 

\paragraph{Q2: What if the class proportion is 0\%?}

In the \emph{Zero-Occur} part of Table~\ref{tab:main}, we also include the discussed extreme situation when the raw corpus does not contain any text falling in the target minority class. A dramatic drop appears in the performance of text mining as there is no ground truth that any miner can get. The data synthesis and text grafting methods are robust to this change as they do not require the existence of ground truth examples. Thus, text grafting is verified to be applicable to raw corpus without the target minority class. Thus, text grafting can be based on a small subset of the corpus which might not contain the target minority class to boost efficiency. 

\begin{figure}
    \centering
    \scalebox{1.0}{\includegraphics[width=\linewidth]{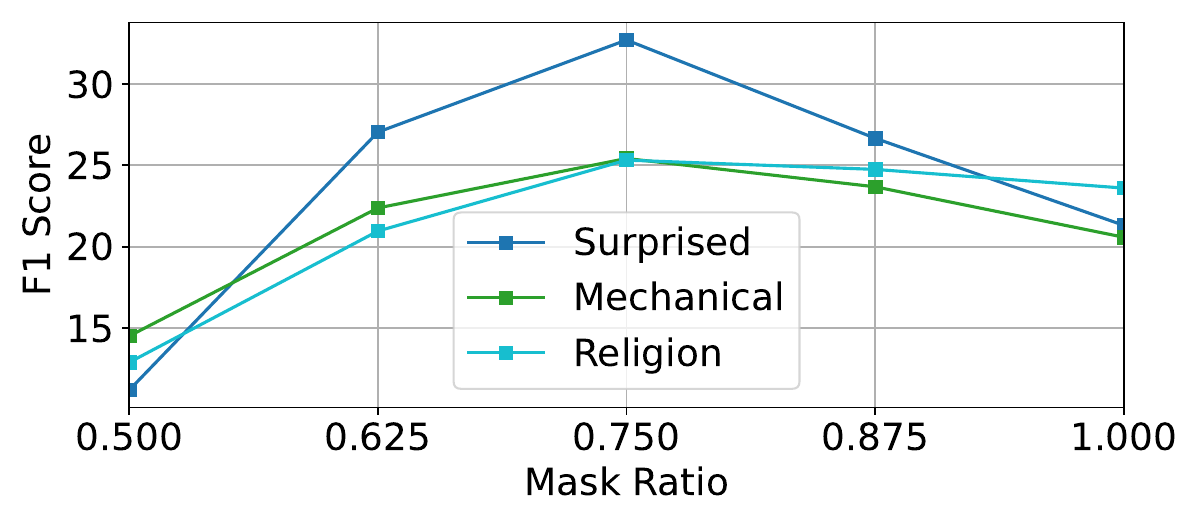}}
    \caption{Analysis of the effect of mask ratio.}
    \label{fig:mask_ratio}
\end{figure}

\paragraph{Q3: How are grafted texts ``near-distribution''?}

\begin{figure}
    \centering
    \scalebox{1.0}{\includegraphics[width=\linewidth]{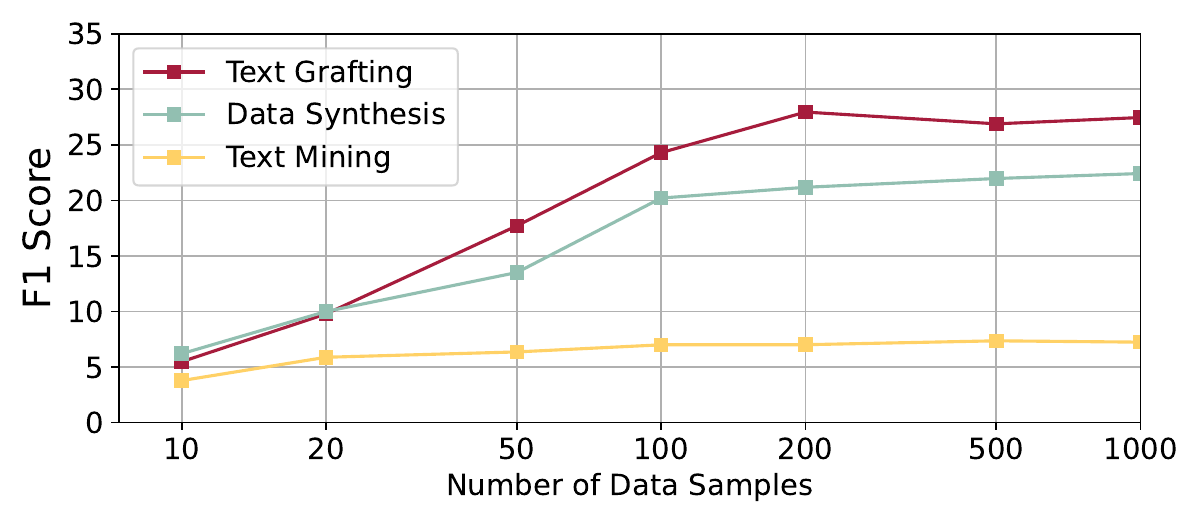}}
    \caption{Analysis of the effect of data number.}
    \label{fig:template_num}
\end{figure}

\begin{figure*}
    \centering
    \scalebox{1.0}{\includegraphics[width=\linewidth]{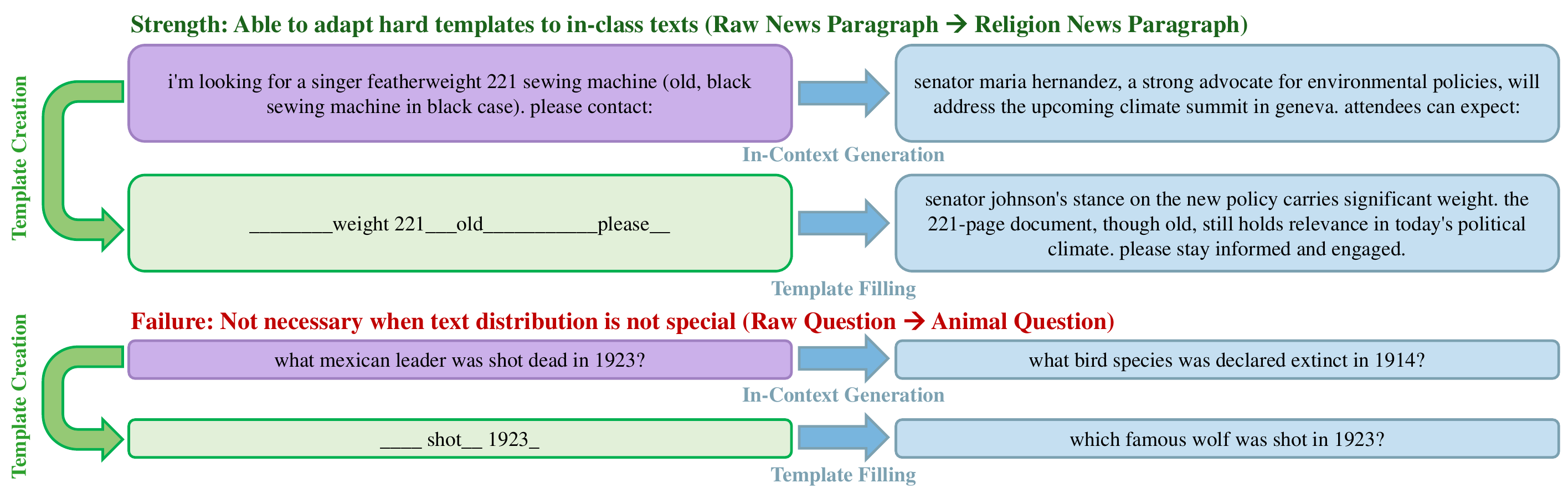}}
    \vspace{-8mm}
    \caption{A case study on the strength and possible failure of text grafting.}
    \label{fig:case}
    \vspace{-5mm}
\end{figure*}

In Figure~\ref{fig:distribution}, we apply semantic text embeddings \cite{simcse} to represent the texts mined or synthesized by different methods. These embeddings are then reduced to 2-dimension by principal component analysis \cite{pca} for visualization. We use the ``Optimism'' class of the TweetEval benchmark and compare the most competitive methods (Debiased Seed Word, Incubator, Text Grafting) of different frameworks. We can observe that text mining only discovers a limited proportion of in-class texts. The synthesized texts fall into a very different domain from the raw corpus, which fine-tunes an out-of-domain classifier with limited generalizability. In contrast, the grafted texts are much more near-distribution, contributing to the performance of the fine-tuned classifier. 

\paragraph{Q4: Is Negative Data Synthesis Necessary?}

For data synthesis-based methods, the synthesis of negative data is an essential stage in the pipeline, which doubles the calls for LLM to synthesize texts. In text grafting, we efficiently use the raw texts as the negative examples. Thus, we explore the necessity of negative synthesis by evaluating the performance of data synthesis (In-Context Generation) and text grafting with or without negative data synthesis with the results presented in Figure~\ref{fig:negative_synthesis}. 

Based on the results, we observe negative data synthesis is very necessary to pure data synthesis as the performance drops dramatically by removing this stage. In contrast, text grafting without negative data synthesis works even better, indicating that our text grafting can work more efficiently by reducing the effort to call LLM at double times. We attribute this efficiency to the near-distribution property of the grafted texts, which makes the discrimination between them and the original raw texts no longer degrade to the classifying of text sources~\cite{detectgpt}.

\paragraph{Q5: What mask ratio to choose?}

In Figure~\ref{fig:mask_ratio}, we analyze the mask ratio used in text grafting. Within the considered set of mask ratios, $\{0.5, 0.625, 0.75, 0.875, 1.0\}$, the best-performing ratio is $0.75$ among different datasets, the same as the setup in our experiments. We can also observe a trend of performance decrease when the mask ratio becomes away from $0.75$. This indicates a too-high masking ratio will make the synthesized text deviate from the domain of raw corpus ($100\%$ leads to in-context generation). On the other hand, a too-low mask ratio will limit the synthesizer to generate in-class texts, which might cause more severe performance drops.

\paragraph{Q6: How many templates to mine?}

In Figure~\ref{fig:template_num}, we further analyze the necessary number of templates to train a strong classifier, which can guide the efficient application of text grafting. The result of the ``surprised'' class shows about $200$ samples can reach the best performance, which results in about $\$0.2$ budget for each class~\cite{openai_hello_gpt4o}. 

We also present how the efficiency of text mining (Debiased Seed Word) and data synthesis (In-Context Generation) is affected by sample numbers. Text mining cannot fine-tune a well-performing classifier due to severe noise in minority class mining. Data synthesis shows a similar scaling trend as text grafting but generally underperforms text grafting. 
\section{Case Study}

In Figure~\ref{fig:case}, we depict workflows of text grafting in comparison with in-context generation to illustrate the strength of grafting and possible failure.

\paragraph{Strength} of text grafting is the ability of state-of-the-art LLMs to fill in hard templates as shown in the first case. While the template is not easy to be grafted into the target ``Politics'' class, the LLM comes up with the methodology to synthesize such a text. The text is also more similar in writing style to the original text than the in-context generation, which depicts the benefit from text grafting. 

\paragraph{Failure} of text grafting can happen when the corpus does not have a writing style very far from the way that LLMs can imitate. As shown in the second case, the LLM can synthesize the animal question without the intermediate template on the TREC corpus~\cite{trec}, which reduces the necessity of text grafting. The XWS-TC of the minority class ``Animal'' on this corpus also shows a similar performance between data synthesis (F1 Score = $53.88$) and text grafting (F1 Score = $53.46$), which again emphasizes ``near-distribution'' to be an essential motivation to use text grafting. 
\section{Conclusion and Future Work}

We introduced text grafting, a technique to generate in-distribution texts for minority classes using LLMs. By mining high-potential masked templates from the raw corpus and filling them with state-of-the-art LLMs, we achieve significant improvements in classifier performance on minority classes. Our analysis and case studies demonstrate the effectiveness of text grafting in enhancing text synthesis for minority classes. Future work will concentrate on improving the precision of template mining and the extension of text grafting to other tasks like information extraction. 
\section*{Limitation}

Despite the presented strengths in the paper, there are still several limitations in the text grafting pipeline. As a hybrid method, text grafting requires a large raw corpus more than data synthesis and LLM calls more than text mining. Other limitations of text grafting also succeed from text mining and data synthesis, such as the dependency on LLM ability (for mining and synthesis). Thus, the application scope for text grafting depends on how LLM comprehends the class name semantics. The performance of different classes might also be biased to the LLM ability in different classes. 

\bibliography{custom}

\clearpage

\appendix

\end{document}